\documentclass{article}

\usepackage{microtype}
\usepackage{graphicx}
\usepackage{booktabs} 

\usepackage{hyperref}

\def\isaccepted{1} 

\usepackage{icml2025}

\usepackage{amsmath}
\usepackage{amssymb}
\usepackage{mathtools}
\usepackage{amsthm}

\usepackage{lipsum}

\usepackage[capitalize,noabbrev]{cleveref}

\theoremstyle{plain}

\theoremstyle{definition}

\theoremstyle{remark}

\usepackage[textsize=tiny]{todonotes}
\usepackage{wrapfig}
\usepackage{array}
\usepackage{tabularx}
\usepackage{multirow} 
\usepackage{subcaption}
\usepackage{caption}
\usepackage{float}
\usepackage{makecell}
\usepackage{pifont}

\begin{document}

\twocolumn[

\icmltitle{\textsc{FlowTS}: Time Series Generation via Rectified Flow}

\icmlsetsymbol{equal}{*}
\icmlsetsymbol{co}{\dag}

\begin{icmlauthorlist}
\icmlauthor{Yang Hu}{equal,west}
\icmlauthor{Xiao Wang}{equal,uw}
\icmlauthor{Zezhen DING}{equal,hkust}
\icmlauthor{Lirong Wu}{west}
\icmlauthor{Huatian Zhang}{ustc}
\icmlauthor{Stan Z. Li}{west}
\icmlauthor{Sheng Wang}{uw}
\icmlauthor{Jiheng Zhang}{hkust}
\icmlauthor{Ziyun Li}{co,kth}
\icmlauthor{Tianlong Chen}{co,unc}
\end{icmlauthorlist}


\icmlaffiliation{unc}{Department of Computer Science, University of North Carolina at Chapel Hill}
\icmlaffiliation{uw}{University of Washington}
\icmlaffiliation{west}{Westlake University}
\icmlaffiliation{ustc}{University of Science and Technology of China}
\icmlaffiliation{hkust}{The Hong Kong University of Science and Technology}
\icmlaffiliation{kth}{KTH Royal Institute of Technology}

\icmlcorrespondingauthor{Tianlong Chen}{tianlong@cs.unc.edu}
\icmlcorrespondingauthor{Ziyun Li}{liziyun2014@gmail.com}
\icmlkeywords{Your Keywords Here}

\vskip 0.3in
]



\printAffiliationsAndNotice{\icmlEqualContribution} 

\begin{abstract}

Diffusion-based models have significant achievements in time series generation but suffer from \textit{inefficient computation}: solving high-dimensional ODEs/SDEs via iterative numerical solvers demands hundreds to thousands of drift function evaluations per sample, incurring prohibitive costs. 
To resolve this, we propose \textsc{FlowTS}, an ODE-based model that leverages rectified flow with \textit{straight-line transport} in probability space. 
By learning geodesic paths between distributions, \textsc{FlowTS} achieves computational efficiency through exact linear trajectory simulation, accelerating training and generation while improving performances.
We further introduce an adaptive sampling strategy inspired by the exploration-exploitation trade-off, balancing noise adaptation and precision. 
Notably, \textsc{FlowTS} enables seamless adaptation from unconditional to conditional generation without retraining, ensuring efficient real-world deployment.
Also, to enhance generation authenticity, \textsc{FlowTS} integrates trend and seasonality decomposition, attention registers (for global context aggregation), and Rotary Position Embedding (RoPE) (for position information).
For unconditional setting, extensive experiments demonstrate that \textsc{FlowTS} achieves state-of-the-art performance, with context FID scores of 0.019 and 0.011 on Stock and ETTh datasets (prev. best: 0.067, 0.061).
For conditional setting, we have achieved superior performance in solar forecasting (MSE 213, prev. best: 375) and MuJoCo imputation tasks (MSE 7e-5, prev. best 2.7e-4). The code is available at 
\url{https://github.com/UNITES-Lab/FlowTS}.
\end{abstract}

\begin{figure}
    \centering
    \includegraphics[width=0.7\linewidth]{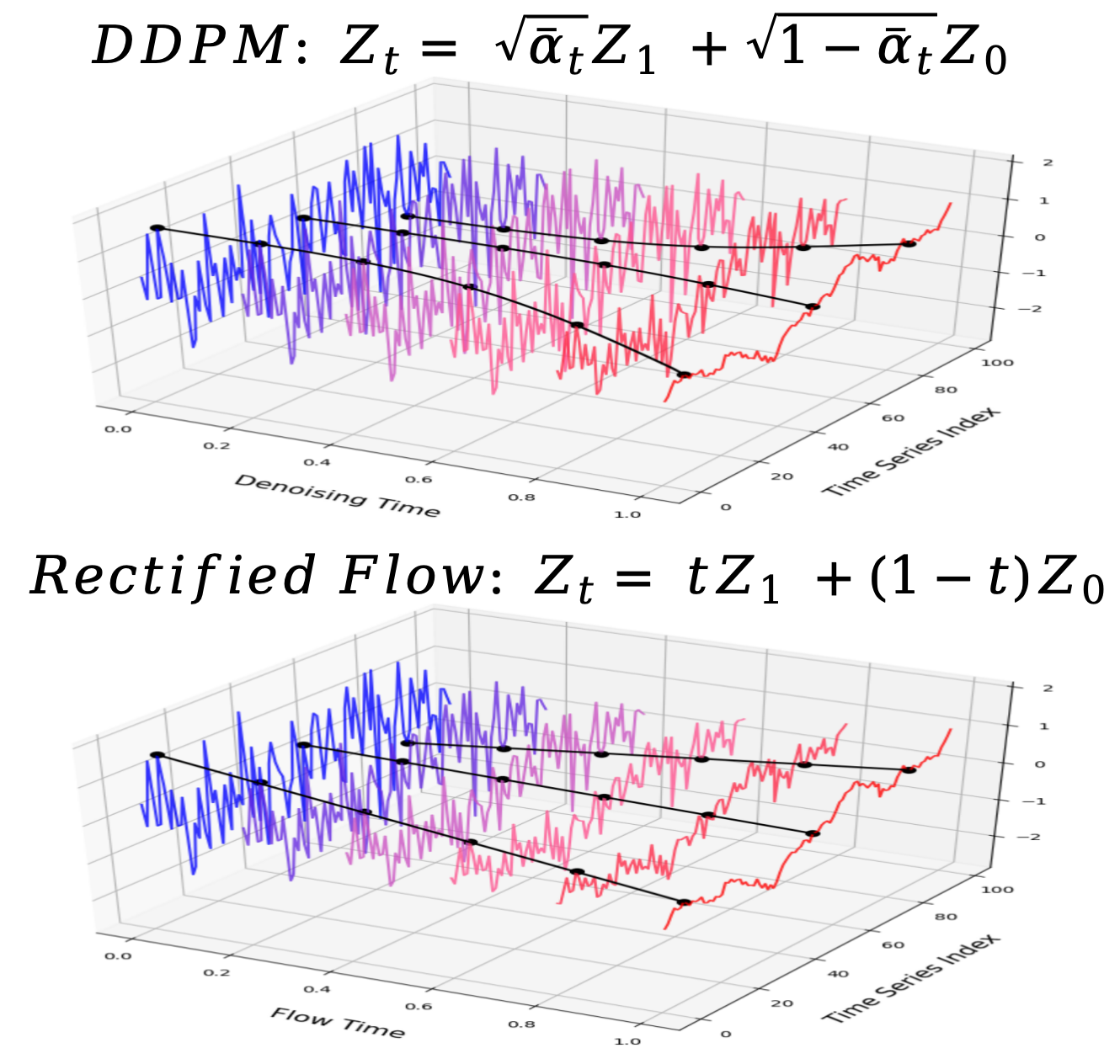}
    \caption{\textbf{Process evolution comparison} {The evolution processes from initial state $Z_0\sim\pi_0$ (blue) to final state $Z_1\sim\pi_1$ (red) are visualized for both models.  The intermediate states (purple) illustrate the continuous evolution between the two endpoints. Rectified Flow exhibits a linear transformation path, while DDPM demonstrates a curved trajectory from $Z_0$ to $Z_1$.}}
    \label{fig:pipeline002}
    \vspace{0mm}
\end{figure}

\vspace{-8mm}
\section{Introduction}
Recent years have seen significant advancements in time series generation, including VAE-based approaches~\citep{desai2021timevaevariationalautoencodermultivariate,xu2020cotgangeneratingsequentialdata}, GAN-based methods, and diffusion-based techniques~\citep{kong2021diffwaveversatilediffusionmodel,tashiro2021csdi}. 
Among these, diffusion-based methods have demonstrated leading performance by explicitly modeling the iterative denoising process, enabling high-quality reconstructions. 
Despite these strengths, a critical limitation persists: \textit{inefficient computation}. 
Generating a single sample requires solving high-dimensional Ordinary/Stochastic Differential Equations (ODEs/SDEs) through iterative numerical solvers, necessitating hundreds to thousands of evaluations of the parameterized drift function, a prohibitive cost for real-world deployment. 
Furthermore, optimizing these models demands extensive hyperparameter searches in complex design spaces \citep{karras2022elucidating}, complicating their adoption even in resource-rich settings.

To address this limitation, we propose \textsc{FlowTS}, an ODE-based generative model designed for highly efficient time series generation. 
Unlike iterative denoising diffusion methods, \textsc{FlowTS} learns to transport the data distribution $\pi_0$ to the target $\pi_1$ via straight-line paths in the probability space. 
The process evolution comparison is demonstrated in Figure~\ref{fig:pipeline002}. 
By bridging the gap between one-step and continuous-time models, \textsc{FlowTS} unifies two key advantages: 
\textit{i)} theoretical optimality through transport-cost-minimizing geodesics; 
\textit{(ii)} computational efficiency via exact simulation along straight trajectories, eliminating iterative steps.
This design accelerates both training and generation while stabilizing the training process by reducing numerical error.
Furthermore, by employing straight-line paths and an unconstrained least-squares procedure for its rectified flow \citep{liu2022flow}, \textsc{FlowTS} avoids the instability issues common to GANs.
We also propose a simple yet highly effective adaptive sampling strategy, inspired by the exploration-exploitation trade-off \cite{sutton2018reinforcement}, which optimally balances an early focus on accommodating higher noise levels with a subsequent shift toward denser and more precise sampling.
Notably, \textsc{FlowTS}’s novel inference mechanism allows a model trained in unconditional settings to seamlessly adapt to conditional tasks without retraining.

%

Another key challenge for time series generation is to ensure its authenticity. 
Following previous practice in diffusion models \citep{yuan2024diffusionts}, we also incorporate trend and seasonality components to explicitly capture periodic and long-term patterns.
However, such explicit decomposition often neglects unobserved global dependencies (e.g., latent interactions). 
To bridge this gap, we exploit attention register tokens\cite{darcet2023vision} to aggregate global context across series and Rotary Position Embedding (RoPE) \cite{su2024roformer} to encode positional information via rotation matrices. 
The extensive experiments validated \textsc{FlowTS} can effectively capture both local and global dependencies through these designs, significantly enhancing the authenticity of generated time series.


In summary, this paper has the following contributions:
\begin{itemize}
\vspace{-2mm}
    \item We introduce \textsc{FlowTS}, a pioneering approach that utilizes rectified flow with ODE-based straight-line transport for time series generation. This design achieves highly efficient generation, and an adaptive sampling strategy further balances exploration and exploitation for improved sample quality.
    \vspace{-1mm}
    \item We explicitly model global information and positional structure using attention register and RoPE, enhancing the authenticity of generated time series.
    \vspace{-1mm}
    \item \textsc{FlowTS} achieves state-of-the-art performance, {with context FID scores of \textit{{0.019}} and \textit{{0.011}} on Stock and ETTh datasets} (prev. best: 0.067, 0.061). {In solar forecasting, it lowers MSE to 213, a \textit{43.2\%} improvement over the previous best of 375}.

\end{itemize}

\section{Related Work}

\subsection{Time Series Generation}

\textbf{VAE \& GAN-based Models}
VAE~\citep{desai2021timevaevariationalautoencodermultivariate} and GAN~\citep{goodfellow2014generative}. TimeVAE~\citep{desai2021timevaevariationalautoencodermultivariate} introduces a hierarchical latent structure to better capture temporal dependencies, while TimeGAN~\citep{yoon2019time} combines adversarial training with supervised learning to preserve both temporal dynamics and feature distributions. CotGAN~\citep{xu2020cotgangeneratingsequentialdata} further improves GAN-based generation by incorporating optimal transport theory to stabilize training.

\textbf{Diffusion-based Models}
Diffusion models, particularly DDPMs ~\citep{ho2020denoising}, emerge as a powerful generative paradigm, offering superior perceptual quality over GANs while avoiding training instability and mode collapse.
In time series, Diffwave ~\citep{kong2020diffwave} pioneers the application of diffusion models to time-varying data through audio synthesis.
For conditional generation, TimeGrad ~\citep{rasul2021autoregressive} employs autoregressive prediction with RNN guidance, while CSDI ~\citep{tashiro2021csdi} and SSSD ~\citep{alcaraz2022diffusionbased} extend the framework to handle irregular time series and missing data imputation using self-supervised masking strategies.
For unconditional generation, TimeDiff ~\citep{shen2023non} adopts a non-autoregressive approach to mitigate boundary disharmony, while ~\cite{lim2023regulartimeseriesgenerationusing} leverages RNNs within Score-based Generative Models (SGMs) \cite{song2021scorebased} for regular series generation.
More recently, TSDiff ~\cite{kollovieh2024predict} introduces self-guided generation with structured state space models, while Diffusion-TS ~\citep{yuan2024diffusionts} enhances temporal modeling with specialized architectures. 

%

\subsection{{Flow Matching}}
Flow matching ~\citep{liu2022flow,lipman2022flow} has emerged as a promising alternative to diffusion models. 
While diffusion models learn to gradually denoise data through a stochastic process, flow matching directly learns the velocity field of a continuous-time transformation, providing deterministic paths with minimal steps. 
Flow matching achieves remarkable results across various domains including video generation ~\citep{klingai2024}, image generation ~\citep{esser2024scalingrectifiedflowtransformers}, point cloud generation ~\citep{wu2023fast}, protein design ~\citep{campbell2024generative,jing2024alphafold}, human motion synthesis ~\citep{hu2023motion}, and speech synthesis ~\citep{mehta2024matcha}. 
Recent work  CFM-TS ~\citep{tamir2024conditional} shows promising results in applying flow matching to Neural ODEs ~\citep{chen2018neural} for time series modeling through conditional probability paths, though its application to time series generation remains unexplored. 

\section{\textsc{FlowTS}}
In this section, we first define the problem statement, outlining the objectives. 
Next, we present the core components of \textsc{FlowTS}, including Rectified Flow-based framework, adaptive sampling strategy, and model architecture.

\subsection{Problem Statement}
\label{sec:problem_statement}
\textbf{Unconditional Generation}
Unconditional time series generation aims to produce sequential data without partially observed data, where the model learns temporal patterns from a training set and generates new sequences following the same distribution. 
Let $X_{1:\ell} = (x_1, \dots, x_\ell) \in \mathbb{R}^{\ell \times d}$ denote a time series of length $\ell$, where $d$ is the number of observed dimensions. 
The problem is defined as follows:
\begin{align}
    \begin{aligned}
    \text{Input:} & \quad Z_0 \sim \pi_0; \\
    & \quad \text{ where } Z_0 \in \mathbb{R}^{\ell \times d}, \pi_0 = \mathcal{N}(0, I). \\
    \text{Output:} & \quad \hat{X}_{1:\ell} = G(Z_0) \in \mathbb{R}^{\ell \times d}; \\
    & \quad G \text{ maps } Z_0 \text{ to the target distribution.}
\end{aligned}
\end{align}
We define $Z_0$ as the initial state, and the generative model \(G\) can be implemented using GANs~\citep{yoon2019time}, VAEs~\citep{desai2021timevaevariationalautoencodermultivariate}, or diffusion models~\citep{tashiro2021csdi,yuan2024diffusionts}, which effectively capture complex temporal dependencies. During training, \(G\) is optimized to minimize the discrepancy between the generated \(\hat{X}_{1:\ell}\) and the target \(X_{1:\ell}\), ensuring high fidelity and diversity in the generated sequences.

\textbf{Conditional Generation}
Conditional time series generation produces sequences by leveraging partially observed data $y$ as context. The generated sequence includes both observed and predicted segments. Formally:
\begin{align}
    \begin{aligned}
\text{Input:} & \quad Z_0 \sim \pi_0, \quad y \in \mathbb{R}^{m \times d} ; \\
& y \text{ is the observed time series with length $m$}. \\
\text{Output:} & \quad \hat{X}_{1:\ell} = G(Z_0; y) \in \mathbb{R}^{\ell \times d}; \\
&  G \text{ maps } Z_0 \text{ to the target distribution given } y.
\end{aligned}
\vspace{-2mm}
\end{align}
%
Conditional time series generation can be divided into two main tasks:  
\textit{i)} \textbf{Forecasting}: \( G \) predicts future values from past observations, given \( y = (x_1, x_2, \ldots, x_m) \) represents observed values up to time \( m \), and the goal is to predict values for \( m+1, m+2, \ldots, \ell \). 
\textit{ii)} \textbf{Imputation}: \( G \) fills unobserved series, where \( y \) represents \( m \) observed series within the range \( 1 \) to \( \ell \).
The key difference between forecasting and imputation lies in the position of known observations.

\begin{figure}[htbp]
    \centering
    \includegraphics[width=1\linewidth]{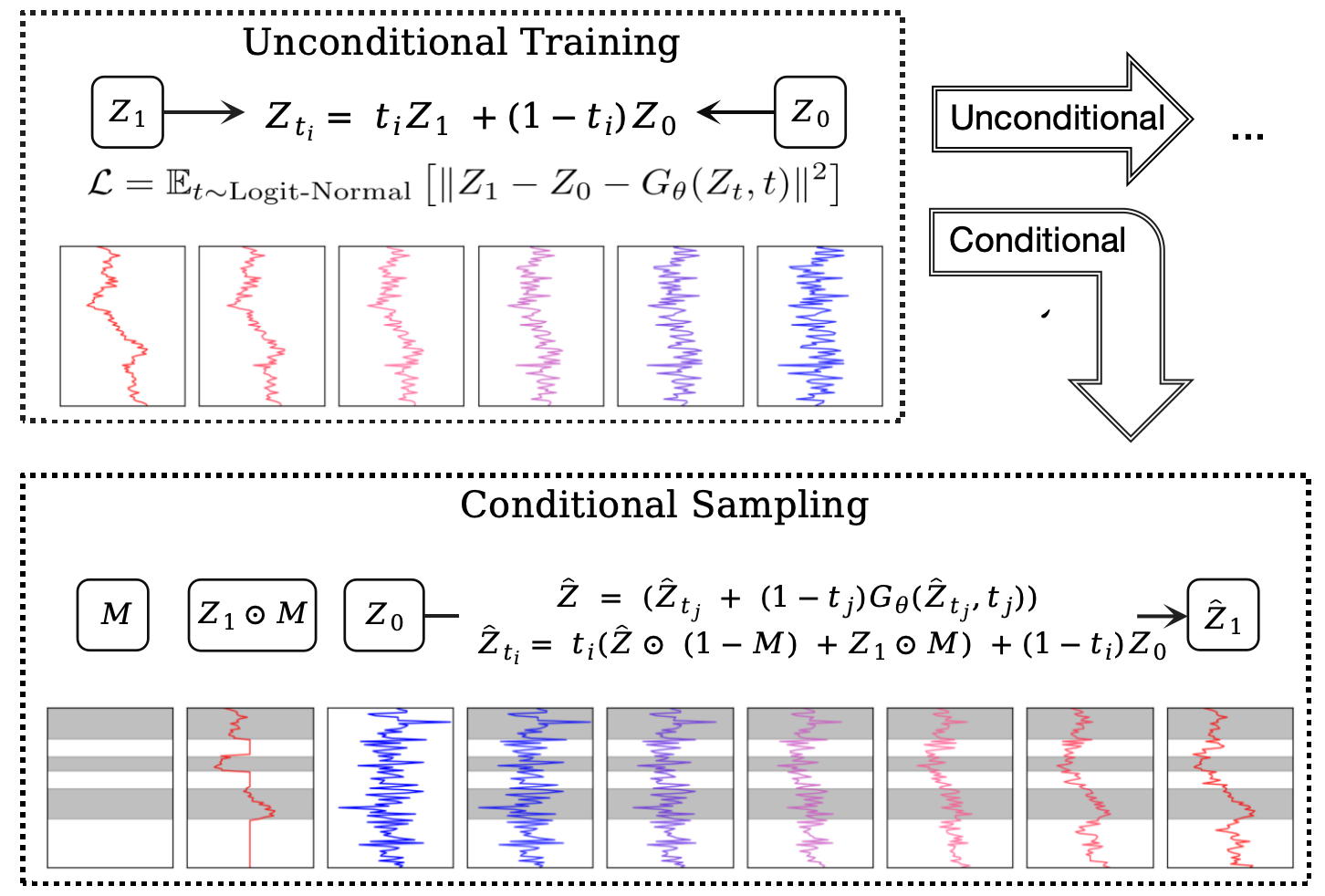}
    \caption{\textbf{\textsc{FlowTS} pipeline.} 
    \textsc{FlowTS} is trained to learn an ODE to transport samples from $Z_0$ to $Z_1$ via a linear path. 
    After training, through adaptive sampling, \textsc{FlowTS} can be applied to both unconditional and conditional generation. 
    The details of sampling are included in Algorithm \ref{alg:fm-ts-unconditional-generation} and \ref{alg:condition}, respectively.}
    \label{fig:pipeline001}
\end{figure}

\subsection{Rectified Flow for Time Series Generation}
To tackle the efficiency challenges in time series generation, we propose \textsc{FlowTS}, leveraging rectified flow as the generative model $G$.
Rectified flow~\citep{liu2022flow} learns an ODE to transport samples between two distributions, \(\pi_0\) and \(\pi_1\).
Therefore, the above problem can be formulated as learning a transport map \(G: \mathbb{R}^{\ell \times d} \rightarrow \mathbb{R}^{\ell \times d}\), such that \(Z_1 := G(Z_0)\) follows the target distribution \(\pi_1\) when \(Z_0 \sim \pi_0\).
By learning \(G\), samples from \(\pi_0\) are transformed to match the target distribution \(\pi_1\), enabling both unconditional and conditional time series generation.

The rectified flow induced from \((Z_0, Z_1)\) is defined as an ODE over time \(t \in [0,1]\):
\begin{equation}
\frac{dZ_t}{dt} = v(Z_t, t), \quad \text{where } t \in [0,1], \, Z_t \in \mathbb{R}^{\ell \times d},
\end{equation}
\(Z_t \in \mathbb{R}^{\ell \times d}\) represents the state at time \(t\), \(v: \mathbb{R}^{\ell \times d} \rightarrow \mathbb{R}^{\ell \times d}\) is the drift function designed to drive the flow along the direction of the linear path \((Z_1 - Z_0)\), ensuring the transformation follows this trajectory as closely as possible.
This can be achieved by solving a least squares regression problem:
\begin{equation}
\min_v \int_0^1 \mathbb{E}\left[\|Z_1 - Z_0 - v(Z_t, t)\|^2\right] dt,
\end{equation}
where $Z_t$ is a linear interpolation between $Z_0$ and $Z_1$, $Z_t = tZ_1 + (1-t)Z_0$, and $v$ is expected to learn with the neural network $G$.
Therefore, model $G$ can be optimized by predicting the direction vector between ($Z_1-Z_0$) via the following loss function:
\begin{equation}
    \mathcal{L} = \mathbb{E}_{t \sim \text{Logit-Normal}}[\| (Z_1 - Z_0)-G(Z_t, t)\|^2],
    \label{eq:overall_loss}
\end{equation}
where $G$ is the model used in \textsc{FlowTS} to learn the drift force $v$. For each sample, $t$ is randomly drawn from a Logit-Normal distribution~\citep{esser2024scalingrectifiedflowtransformers}, while $Z_1$ is sampled from the target time series distribution $\pi_1$, and $Z_0$ is sampled from the standard normal distribution $\pi_0$.

The overall framework is illustrated in Fig.~\ref{fig:pipeline001}. 
The unconditional time series generation model \( G \) is trained using the loss in Eq.~\ref{eq:overall_loss}, where it takes \( Z_t \) and \( t \) as inputs to predict the drift force \( v \) between \( Z_0 \) and \( Z_1 \). 
Once trained, the unconditional model can be directly applied to conditional generation tasks without requiring task-specific training for a separate conditional model.
Instead, it can be achieved by a simple adaptive sampling, illustrated in Section~\ref{sec:adative}.

\begin{algorithm}
\caption{FlowTS unconditional generation}
\label{alg:fm-ts-unconditional-generation}
\begin{algorithmic}[1]
\REQUIRE Sampling iterations $N$, adaptive parameter $k$, trained flow matching model $G_\theta$
\STATE $\hat{\mathbf{Z}}0 \sim \mathcal{N}(0, \mathbf{I})$ \hfill $\triangleright${Sample initial noise}
\STATE $t_0 = 0$ \hfill $\triangleright${Initialize time step}
\FOR{$i=0$ to $N-1$}
\STATE$t_{i+1} = ((i+1) / N)^{k}$ \hfill $\triangleright${Get next adaptive time step}
\STATE $\hat{\mathbf{v}}_{t{i}} = G_{\theta}(\mathbf{Z}_{t_{i}}, t_{i})$ \hfill $\triangleright${Get velocity from model}
\STATE $\hat{\mathbf{Z}}_{t_{i+1}} = \hat{\mathbf{Z}}_{t_{i}} + (t_{i+1} - t_{i}) \hat{\mathbf{v}}_{t_{i}}$ \hfill $\triangleright${One Euler step}
\ENDFOR
\STATE \textbf{return} $\hat{\mathbf{Z}}_1$
\end{algorithmic}
\end{algorithm}

\subsection{{Adaptive Sampling}}
\label{sec:adative}
The exploration-exploitation trade-off balances discovering new strategies and leveraging known ones, a key concept in decision-making under uncertainty \cite{sutton2018reinforcement, lattimore2020bandit, jin2018qlearning, agarwal2021policy}.
Inspired by this, we introduce adaptive sampling by using a scaling factor \( t^k \), where \( k \in (0, 1] \) (\( k = 1 \) represents uniform sampling). 
Early in the process, \( t^k \) promotes exploration by encouraging the model to estimate higher noise levels, while in later stages, it focuses on exploitation through denser, smaller sampling steps. 
Furthermore, we observe that as the number of sampling iterations increases, the optimal \( k \) decreases. 
Detailed experiments on the impact of \( k \) are provided in {Section~\ref{sec:fur}}.

For unconditional sampling, we utilize ODE sampling along a straightened trajectory. This method reduces the number of sampling steps, enabling faster inference while maintaining performance, as detailed in Algorithm~\ref{alg:fm-ts-unconditional-generation}.
For conditional sampling, inspired by \cite{song2020score}, we used the observed segment as context to refine the sampling process.
As described in Algorithm~\ref{alg:condition}, the conditional generation includes two main novel designs compared to unconditional setting.
First, it utilized the observed $Z_1$ to refine the $\hat{Z_1}$ estimation (line 5) at every iteration.
Then, it will interpolate the $\hat{Z_t}$ at timestep $t$ given refined $\hat{Z_1}$ (line 6).
After that, it will be same as the unconditional generation to estimate the $\hat{Z}_{t
+1}$ at next timestep $t+1$.
Therefore, our framework enables seamless adaptation from unconditional to conditional generation without retraining.

\begin{algorithm}
\caption{FlowTS conditional generation}
\label{alg:condition}
\begin{algorithmic}[1]
\REQUIRE Target time series $\mathbf{Z_1}$, observation mask $\mathbf{M}$, sampling iterations $N$, adaptive parameter $k$, trained flow matching model $G_\theta$
\STATE $\hat{\mathbf{Z}}_1 \sim \mathcal{N}(0, \mathbf{I})$  \hfill $\triangleright${Initialize with random noise}
\FOR{$i=0$ to $N-1$}
\STATE $t_{i} = (i / N)^{k}$  \hfill $\triangleright${Compute adaptive time step}
\STATE $\mathbf{Z}_0 \sim \mathcal{N}(0, \mathbf{I})$ \hfill 
 $\triangleright${Sample noise}
\STATE $\hat{\mathbf{Z}}_1 = \hat{\mathbf{Z}}_1 \odot (1-\mathbf{M}) + \mathbf{Z}_1 \odot \mathbf{M}$  \hfill $\triangleright${Replace with observed values}
\STATE $\hat{\mathbf{Z}}_{t_{i}} = t_{i} \hat{\mathbf{Z}}_1 + (1-t_{i}) \mathbf{Z}_0$  \hfill $\triangleright${Interpolate}
\STATE $\hat{\mathbf{v}}_{t_{i}} = G_{\theta}(\hat{\mathbf{Z}}_{t_{i}}, t_{i})$  \hfill $\triangleright${Flow matching step}
\STATE $\hat{\mathbf{Z}}_1 = \hat{\mathbf{Z}}_{t_{i}} + (1 - t_{i}) \hat{\mathbf{v}}_{t_{i}}$  \hfill $\triangleright${One Euler step}
\ENDFOR
\STATE \textbf{return} $\hat{\mathbf{Z}}_1$
\end{algorithmic}
\end{algorithm}

\subsection{{Model Architecture}}

\begin{figure}
    \centering
    \includegraphics[width=0.95\linewidth]{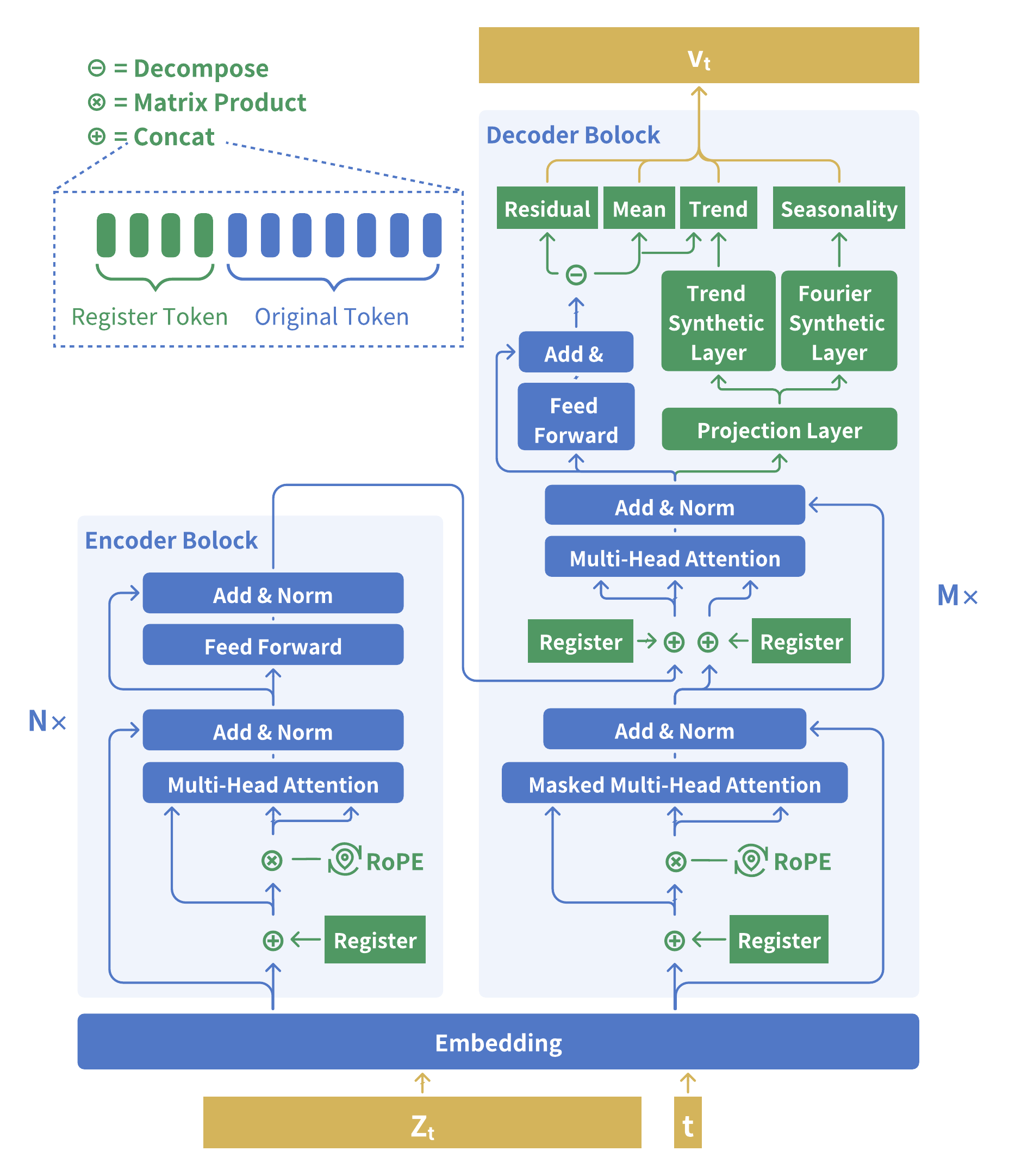}
    \caption{The \textsc{FlowTS} model builds on the standard Transformer architecture. The blue components represent the basic encoder-decoder structure, while the green components highlight novel additions, including attention registers, RoPE, Trend and Fourier Synthetic Layers. The model processes input $Z_t$ and timestep $t$ through $N$ encoder and $M$ decoder blocks to produce the drift force $v_t$ at time $t$.
}
\label{fig:model-architechure}
\end{figure}

\paragraph{Overview}
As shown in Figure~\ref{fig:model-architechure}, \textsc{FlowTS} is based on the encoder-decoder Transformer~\citep{vaswani2017attention}, with blue components as the standard structure and green highlighting novel additions.
Attention register \cite{darcet2023vision} enhances the model's memory capability and information transmission efficiency, enabling efficient capture of global dependencies. 
The integration of Rotary Position Embedding (RoPE) \cite{su2024roformer} provides precise relative positional encoding, enhancing the model's ability to represent temporal relationships effectively.
Inspired by~\cite{yuan2024diffusionts}, our model incorporates trend synthetic layers and fourier synthetic layers to capture the trend and periodic patterns. This design improves interpretability by decomposing the prediction target into residual, mean, trend, and seasonality components.

\textbf{Attention Register}
As shown in {Figure \ref{fig:model-architechure}}, attention registers are learnable tokens that serve as dedicated computational units, distinct from the main sequence tokens~\citep{darcet2023vision}. As persistent memory, these registers capture and maintain global patterns, allowing sequence tokens to focus on local temporal dependencies. This mechanism has proven effective in vision~\citep{darcet2023vision} and language models~\citep{xiao2023streamingllm}. Thus we integrate learnable register tokens into our architecture to capture global information.

\textbf{Rotary Position Embedding (RoPE)}
RoPE ~\citep{su2024roformer} is a position encoding mechanism that enhances the standard Transformer architecture by encoding positional information through rotation matrices. 
We also incorporate RoPE into our time series generation framework for two key advantages: \textit{i)} its natural decay property, which aligns with temporal dependencies in time series data, and \textit{ii)} its flexibility in handling varying sequence lengths.

\section{Experiments}

\begin{table*}[t]
    \caption{Unconditional time series generation benchmark with 24-length} 
    \centering
    \resizebox{0.8\textwidth}{!}{
    \footnotesize
    \newcolumntype{Y}{>{\centering\arraybackslash}X}
    \begin{tabularx}{\textwidth}{@{}ll*{6}{Y}@{}}
    \toprule
    \textbf{Metric} & \textbf{Methods} & \textbf{Sines} & \textbf{Stocks} & \textbf{ETTh} & \textbf{MuJoCo} & \textbf{Energy} & \textbf{fMRI} \\
    \midrule
    \multirow{7}{*}{\makecell[l]{Discriminative  \\ (Lower Better)}}
    & \textbf{\textsc{FlowTS}} & \textbf{0.005±.005} & \textbf{0.019±.013} & \textbf{0.011±.015} & \textbf{0.005±.005} & \textbf{0.053±.010} & \textbf{0.106±.018} \\
    & Diffusion-TS & 0.006±.007 & 0.067±.015 & 0.061±.009 & 0.008±.002 & 0.122±.003 & 0.167±.023 \\
    & TimeGAN & 0.011±.008 & 0.102±.021 & 0.114±.055 & 0.238±.068 & 0.236±.012 & 0.484±.042 \\
    & TimeVAE & 0.041±.044 & 0.145±.120 & 0.209±.058 & 0.230±.102 & 0.499±.000 & 0.476±.044 \\
    & Diffwave & 0.017±.008 & 0.232±.061 & 0.190±.008 & 0.203±.096 & 0.493±.004 & 0.402±.029 \\
    & DiffTime & 0.013±.006 & 0.097±.016 & 0.100±.007 & 0.154±.045 & 0.445±.004 & 0.245±.051 \\
    & Cot-GAN & 0.254±.137 & 0.230±.016 & 0.325±.099 & 0.426±.022 & 0.498±.002 & 0.492±.018 \\
    \midrule
    \multirow{8}{*}{\makecell[l]{Predictive  \\ (Lower Better)}}
    & \textbf{\textsc{FlowTS}} & \textbf{0.092±.000} & \textbf{0.036±.000} & \textbf{0.118±.005} & 0.008±.001 & \textbf{0.250±.000} & \textbf{0.099±.000} \\
    & Diffusion-TS & 0.093±.000 & 0.036±.000 & 0.119±.002 & \textbf{0.007±.000} & \textbf{0.250±.000} & \textbf{0.099±.000} \\
    & TimeGAN & 0.093±.019 & 0.038±.001 & 0.124±.001 & 0.025±.003 & 0.273±.004 & 0.126±.002 \\
    & TimeVAE & 0.093±.000 & 0.039±.000 & 0.126±.004 & 0.012±.002 & 0.292±.000 & 0.113±.003 \\
    & Diffwave & 0.093±.000 & 0.047±.000 & 0.130±.001 & 0.013±.000 & 0.251±.000 & 0.101±.000 \\
    & DiffTime & 0.093±.000 & 0.038±.001 & 0.121±.004 & 0.010±.001 & 0.252±.000 & 0.100±.000 \\
    & Cot-GAN & 0.100±.000 & 0.047±.001 & 0.129±.000 & 0.068±.009 & 0.259±.000 & 0.185±.003 \\
    & Original & 0.094±.001 & 0.036±.001 & 0.121±.005 & 0.007±.001 & 0.250±.003 & 0.090±.001 \\
    \midrule 
    \multirow{7}{*}{\makecell[l]{Context-FID  \\ (Lower Better)}} 
    & \textbf{\textsc{FlowTS}} & \textbf{0.002±.000} & \textbf{0.015±.003} & \textbf{0.024±.001} & \textbf{0.009±.000} & \textbf{0.031±.004} & 0.128±.009 \\
    & Diffusion-TS & 0.006±.000 & 0.147±.025 & 0.116±.010 & 0.013±.001 & 0.089±.024 & \textbf{0.105±.006} \\
    & TimeGAN & 0.101±.014 & 0.103±.013 & 0.300±.013 & 0.563±.052 & 0.767±.103 & 1.292±.218 \\
    & TimeVAE & 0.307±.060 & 0.215±.035 & 0.805±.186 & 0.251±.015 & 1.631±.142 & 14.449±.969 \\
    & Diffwave & 0.014±.002 & 0.232±.032 & 0.873±.061 & 0.393±.041 & 1.031±.131 & 0.244±.018 \\
    & DiffTime & 0.006±.001 & 0.236±.074 & 0.299±.044 & 0.188±.028 & 0.279±.045 & 0.340±.015 \\
    & Cot-GAN & 1.337±.068 & 0.408±.086 & 0.980±.071 & 1.094±.079 & 1.039±.028 & 7.813±.550 \\
    \midrule
    \multirow{7}{*}{\makecell[l]{Correlational  \\ (Lower Better)}}
    & \textbf{\textsc{FlowTS}} & 0.015±.006 & 0.012±.011 & \textbf{0.022±.010} & \textbf{0.183±.051} & \textbf{0.650±.201} & \textbf{0.938±.039} \\
    & Diffusion-TS & \textbf{0.015±.004} & \textbf{0.004±.001} & 0.049±.008 & 0.193±.027 & 0.856±.147 & 1.411±.042 \\
    & TimeGAN & 0.045±.010 & 0.063±.005 & 0.210±.006 & 0.886±.039 & 4.010±.104 & 23.502±.039 \\
    & TimeVAE & 0.131±.010 & 0.095±.008 & 0.111±.020 & 0.388±.041 & 1.688±.226 & 17.296±.526 \\
    & Diffwave & 0.022±.005 & 0.030±.020 & 0.175±.006 & 0.579±.018 & 5.001±.154 & 3.927±.049 \\
    & DiffTime & 0.017±.004 & 0.006±.002 & 0.067±.005 & 0.218±.031 & 1.158±.095 & 1.501±.048 \\
    & Cot-GAN & 0.049±.010 & 0.087±.004 & 0.249±.009 & 1.042±.007 & 3.164±.061 & 26.824±.449 \\
    \bottomrule
    \end{tabularx}
    }
    \label{tab:unconditional}
    \end{table*}

\begin{figure}[htbp]
    \centering
    \includegraphics[width=0.48\textwidth]{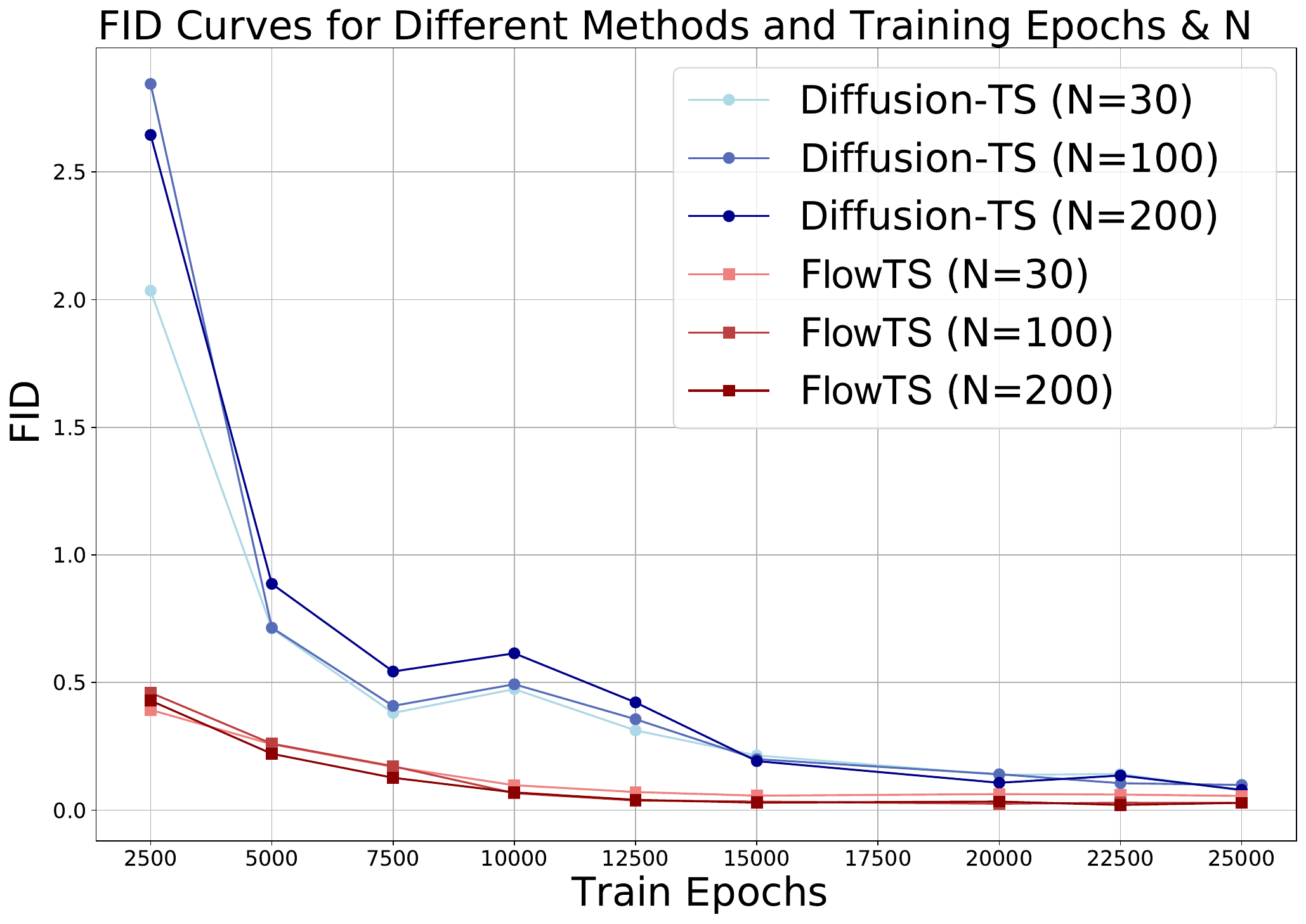}
    \caption{FID curves of \textsc{FlowTS} and Diffusion-TS on {Energy} dataset across training epochs (2,500–25,000) and sampling steps. 
    \textsc{FlowTS} demonstrates superior efficiency, achieving lower FID scores with fewer training and sampling epochs.}
    \label{fig:eff}
\end{figure}

\subsection{Experiment Setup}
\textbf{Datasets}
Our evaluation employs six diverse datasets:
the three real-world datasets include \textit{Stocks}~\footnote{finance.yahoo.com/quote/GOOG/history?p=GOOG} for measuring daily stock price data, \textit{ETTh}~\footnote{github.com/zhouhaoyi/ETDataset}~\citep{haoyietal-informer-2021} for interval electricity transformer data, and \textit{Energy}~\footnote{archive.ics.uci.edu/ml/datasets/Appliances+energy+prediction}
 for UCI appliance energy prediction.
The three simulation data sets include \textit{fMRI}~\footnote{fmrib.ox.ac.uk/datasets/netsim/} for simulated \mbox{blood-oxygen-level-dependent} time series, \textit{Sines}~\footnote{github.com/jsyoon0823/TimeGAN}~\citep{yoon2019time} generated from different frequencies, amplitudes, and phases, and \textit{Mujoco}~\footnote{github.com/google-deepmind/dm\_control} from multivariate simulation of physics.
These datasets capture diverse time series characteristics, including periodicity, dimensionality, and feature correlation, ensuring a comprehensive evaluation of our approach.

\textbf{Metrics}
Following the practices of time series generation~\citep{yuan2024diffusionts}, we evaluate our method using four metrics:  
1) \textbf{Discriminative Score}~\citep{yoon2019time}, which measures distributional similarity between real and synthetic data using a 2-layer LSTM classifier; lower classification error indicates higher-quality synthetic data.  
2) \textbf{Predictive Score}~\citep{yoon2019time}, which assesses the utility of synthetic data for prediction tasks by training a 2-layer LSTM on synthetic data and testing on real data, with performance measured by mean absolute error (MAE).  
3) \textbf{Context-FID}~\citep{jeha2022psa}, which quantifies quality by comparing contextual representations, capturing both distributional similarity and temporal dependencies.  
4) \textbf{Correlational Score}~\citep{liao2020conditional}, which evaluates the preservation of temporal dependencies by comparing cross-correlation matrices, where lower errors indicate better structural fidelity. These metrics provide a comprehensive evaluation of the quality, utility, and temporal consistency of synthetic time series data.

\textbf{Baselines \& Implementation Details}
We evaluate our method for unconditional time series generation by comparing with six baseline models: Diffusion-TS \cite{yuan2024diffusionts}, TimeGAN~\citep{yoon2019time}, TimeVAE~\citep{desai2021timevaevariationalautoencodermultivariate}, Diffwave~\citep{kong2020diffwave} and CotGAN~\citep{xu2020cotgangeneratingsequentialdata} 
For conditional tasks, we also compare our method against CSDI~\citep{tashiro2021csdi} and SSSD~\citep{alcaraz2022diffusionbased}.
For a fair comparison, we closely follow the Diffusion-TS experimental setup, using the same hyperparameters, including batch size, attention heads, learning rate (AdamW), training iterations, etc.
Specifically, 
Specifically, for 24-length, we use 4 attention register tokens with RoPE set to a frequency of 50k. For 256-length, we increase the attention registers to 128 while keeping the RoPE frequency unchanged.

\subsection{Higher Efficiency}

\begin{figure}
    \centering
    \includegraphics[width=0.48\textwidth]{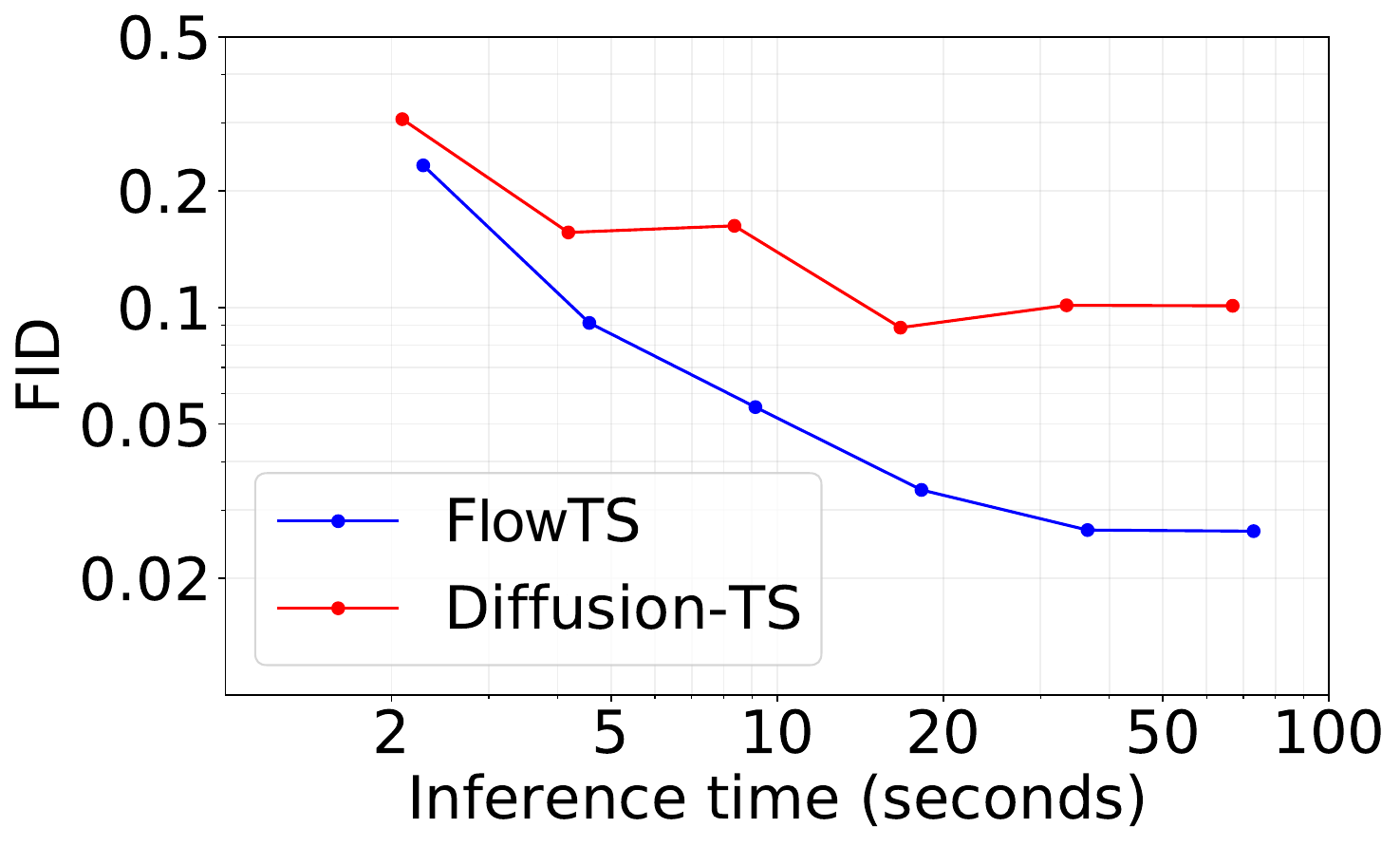}
    \caption{ Context-FID scores with different sampling iterations (1, 2, 4, 8, 16, 32) on Energy, measured against cumulative sampling time.
    \textsc{FlowTS} outperforms Diffusion-TS in both accuracy and sampling speed.}
    \label{timefid}
\end{figure}
\textbf{Training Efficiency}
To evaluate training efficiency, we benchmarked \textsc{FlowTS} and Diffusion-TS across multiple training iterations on the \textit{Energy} dataset. As shown in Figure~\ref{fig:eff}, \textsc{FlowTS} consistently outperforms Diffusion-TS in achieving superior FID scores, with training iterations ranging from 2,500 to 25,000.
Remarkably, \textsc{FlowTS} achieves superior performance with only 30 sampling iterations ($N=30$) after just 2,500 training iterations, whereas Diffusion-TS fails to reach the same level of efficiency even with 200 inference steps ($N=200$) after 10,000 training iterations. 

\textbf{Inference Efficiency}
To assess inference efficiency, we measure the sampling time (seconds) on the \textit{Energy} dataset with varying sampling steps (1, 2, 4, 8, 16, 32) for Diffusion-TS and \textsc{FlowTS}. 
As shown in Figure~\ref{timefid}, \textsc{FlowTS} outperforms Diffusion-TS in both accuracy and sampling speed, demonstrating its efficiency in time series generation.



\subsection{Better Performance}

\begin{table*}[h]
    \caption{Benchmark of Unconditional Long-term Time Series Generation on ETTh Datasets}
    \centering
    \resizebox{0.98\textwidth}{!}{
    \begin{tabular}{@{}ccccccccc@{}}
    \toprule
    \textbf{Metric} & \textbf{Length} & \textbf{\textsc{FlowTS}} & \textbf{Diffusion-TS} & \textbf{TimeGAN} & \textbf{TimeVAE} & \textbf{Diffwave} & \textbf{DiffTime} & \textbf{Cot-GAN} \\
    \midrule

    \multirow{3}{*}{\makecell[c]{Discriminative\\(Lower Better)}} 
    & 64  & \textbf{0.010 ± 0.004} & 0.106 ± 0.048 & 0.227 ± 0.078 & 0.171 ± 0.142 & 0.254 ± 0.074 & 0.150 ± 0.003 & 0.296 ± 0.348 \\
    & 128 & \textbf{0.040 ± 0.012} & 0.144 ± 0.060 & 0.188 ± 0.074 & 0.154 ± 0.087 & 0.274 ± 0.047 & 0.176 ± 0.015 & 0.451 ± 0.080 \\
    & 256 & 0.081 ± 0.022 & \textbf{0.060 ± 0.030} & 0.444 ± 0.056 & 0.178 ± 0.076 & 0.304 ± 0.068 & 0.243 ± 0.005 & 0.461 ± 0.010 \\
    \midrule

    \multirow{3}{*}{\makecell[c]{Predictive\\(Lower Better)}} 
    & 64  & \textbf{0.115 ± 0.005} & 0.116 ± 0.000 & 0.132 ± 0.008 & 0.118 ± 0.004 & 0.133 ± 0.008 & 0.118 ± 0.004 & 0.135 ± 0.003 \\
    & 128 & \textbf{0.104 ± 0.013} & 0.110 ± 0.003 & 0.153 ± 0.014 & 0.113 ± 0.005 & 0.129 ± 0.003 & 0.120 ± 0.008 & 0.126 ± 0.001 \\
    & 256 & \textbf{0.107 ± 0.005} & 0.109 ± 0.013 & 0.220 ± 0.008 & 0.110 ± 0.027 & 0.132 ± 0.001 & 0.118 ± 0.003 & 0.129 ± 0.000 \\
    \midrule

    \multirow{3}{*}{\makecell[c]{Context-FID\\(Lower Better)}} 
    & 64  & \textbf{0.039 ± 0.003} & 0.631 ± 0.058 & 1.130 ± 0.102 & 0.827 ± 0.146 & 1.543 ± 0.153 & 1.279 ± 0.083 & 3.008 ± 0.277 \\
    & 128 & \textbf{0.128 ± 0.007} & 0.787 ± 0.062 & 1.553 ± 0.169 & 1.062 ± 0.134 & 2.354 ± 0.170 & 2.554 ± 0.318 & 2.639 ± 0.427 \\
    & 256 & \textbf{0.302 ± 0.018} & 0.423 ± 0.038 & 5.872 ± 0.208 & 0.826 ± 0.093 & 2.899 ± 0.289 & 3.524 ± 0.830 & 4.075 ± 0.894 \\
    \midrule

    \multirow{3}{*}{\makecell[c]{Correlational\\(Lower Better)}} 
    & 64  & \textbf{0.027 ± 0.015} & 0.082 ± 0.005 & 0.483 ± 0.019 & 0.067 ± 0.006 & 0.186 ± 0.008 & 0.094 ± 0.010 & 0.271 ± 0.007 \\
    & 128 & \textbf{0.030 ± 0.011} & 0.088 ± 0.005 & 0.188 ± 0.006 & 0.054 ± 0.007 & 0.203 ± 0.006 & 0.113 ± 0.012 & 0.176 ± 0.006 \\
    & 256 & \textbf{0.025 ± 0.008} & 0.064 ± 0.007 & 0.522 ± 0.013 & 0.046 ± 0.007 & 0.199 ± 0.003 & 0.135 ± 0.006 & 0.222 ± 0.010 \\
    
    \bottomrule
    \end{tabular}
    }
    \label{tablelong}
\end{table*}

\textbf{Unconditional Sampling}
As shown in Table \ref{tab:unconditional}, \textsc{FlowTS} demonstrates significantly superior performance across all evaluation metrics compared to  all baselines on short-term 24-length time series.
For the Discriminative Score, \textsc{FlowTS} reduces error by up to 50\% compared to Diffusion-TS (e.g., 0.005 vs. 0.006 on Sines). On the Predictive Score, \textsc{FlowTS} matches or slightly improves upon Diffusion-TS (e.g., 0.036 vs. 0.036 on Stocks). For Context-FID, \textsc{FlowTS} achieves up to 33\% lower scores (e.g., 0.009 vs. 0.013 on MuJoCo), while for the Correlational Score, it provides up to 15\% better temporal correlation preservation (e.g., 0.938 vs. 1.411 on fMRI). 
Also, \textsc{FlowTS} achieves superior performance on long-term time series generation across all metrics on the ETTh dataset. For Discriminative Score, it improves by up to 83\% over the second-best method (e.g., 0.010 vs. 0.060 for length 64). For Predictive Score, it consistently outperforms, with up to 6\% better results (e.g., 0.104 vs. 0.110 for length 128). On Context-FID, \textsc{FlowTS} achieves scores up to 88\% lower (e.g., 0.039 vs. 0.423 for length 64), and for Correlational Score, it preserves temporal structure with up to 61\% improvement (e.g., 0.025 vs. 0.064 for length 256). 
These results highlight \textsc{FlowTS} as the most effective method for both short term and long-term time series generation.

 

\begin{figure}[htb]
    \centering
    \includegraphics[width=0.48\textwidth]{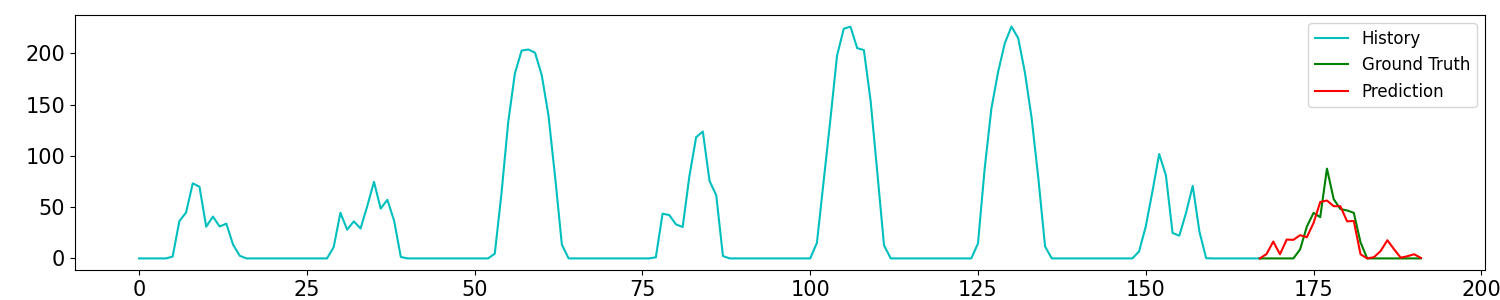}
    \caption{An example of solar forecasting results.}
    \label{solar_demo}
\end{figure}

\textbf{Conditional Sampling}
After validating \textsc{FlowTS} on unconditional time series generation, we further assessed its generalizability for conditional time series generation. Without retraining the model, we utilized a specialized inference algorithm (Algorithm~\ref{alg:condition}) to incorporate observed information during inference. As outlined in Section~\ref{sec:problem_statement}, conditional time series generation involves two key tasks: forecasting and imputation. To demonstrate the effectiveness of \textsc{FlowTS}, we benchmarked it on the \textit{Solar} and \textit{Mujoco} datasets, following the practices in~\citep{alcaraz2022diffusion} and~\citep{tashiro2021csdi}.

\textit{(i)} \textbf{Forecasting:} 
Table~\ref{tab:combined_results} summarizes the forecasting performance on the Solar dataset. With a sequence length of 168, \textsc{FlowTS} achieved a MSE of 213 for the next 24 time points, significantly outperforming Diffusion-TS (375). This demonstrates the superior prediction accuracy and sequence alignment of \textsc{FlowTS}. Figure~\ref{solar_demo} further illustrates the forecasting results, showing how \textsc{FlowTS} effectively captures the incoming peak in the future time series.

\textit{(ii)} \textbf{Imputation:}
We further evaluated \textsc{FlowTS} on the imputation task using the MuJoCo dataset (Table~\ref{tab:combined_results}), following the setup in~\citep{alcaraz2022diffusionbased}. Despite most competing methods being tailored for conditional time series generation, \textsc{FlowTS} consistently outperformed them under varying missing data ratios. Notably, for a 70\% missing rate, it achieved a Mean Squared Error (MSE) of 0.00007, a 74.1\% reduction compared to Diffusion-TS (0.00027), highlighting its superior imputation performance.

\begin{table}[h]
\centering
\small
\setlength{\tabcolsep}{2.2pt} 
\caption{Time series forecasting on Solar and imputation on Mujoco. MSE results of Mujoco are in the order of 1e-3.}
\begin{tabular}{lccc}
\toprule
\multirow{2}{*}{Model} & Solar Forecasting & \multicolumn{2}{c}{Mujoco Imputation} \\
\cmidrule(lr){2-2} \cmidrule(lr){3-4}
 & \text{168} $\rightarrow$ \text{24}
  & Missing(70 \%) & Missing(80 \%) \\
\midrule
GP-copula & 9.8e2 & -- & -- \\
TransMAF & 9.30e2 & -- & -- \\
TLAE & 6.8e2 & -- & -- \\
RNN GRU-D & -- & 11.34 & 14.21 \\
ODE-RNN & -- & 9.86 & 12.09 \\
NeuralCDE & -- & 8.35 & 10.71 \\
Latent-ODE & -- & 3.00 & 2.95 \\
NAOMI & -- & 1.46 & 2.32 \\
NRTSI & -- & 0.63 & 1.22 \\
CSDI & 9.0e2 & 0.24 & 0.61 \\
SSSD & 5.03e2 & 0.59 & 1.00 \\
Diffusion-TS & 3.75e2 & 0.27 & 0.32 \\
\textsc{FlowTS} & \textbf{2.13e2} & \textbf{0.07} & \textbf{0.14} \\
\bottomrule
\end{tabular}
\label{tab:combined_results}
\end{table}

\begin{figure*}[]
    \centering
    \subfloat[MSE with changing $N$ and $k$ on Mujoco dataset imputation tasks, with missing ratio 0.7]{
        \includegraphics[width=0.3\textwidth]{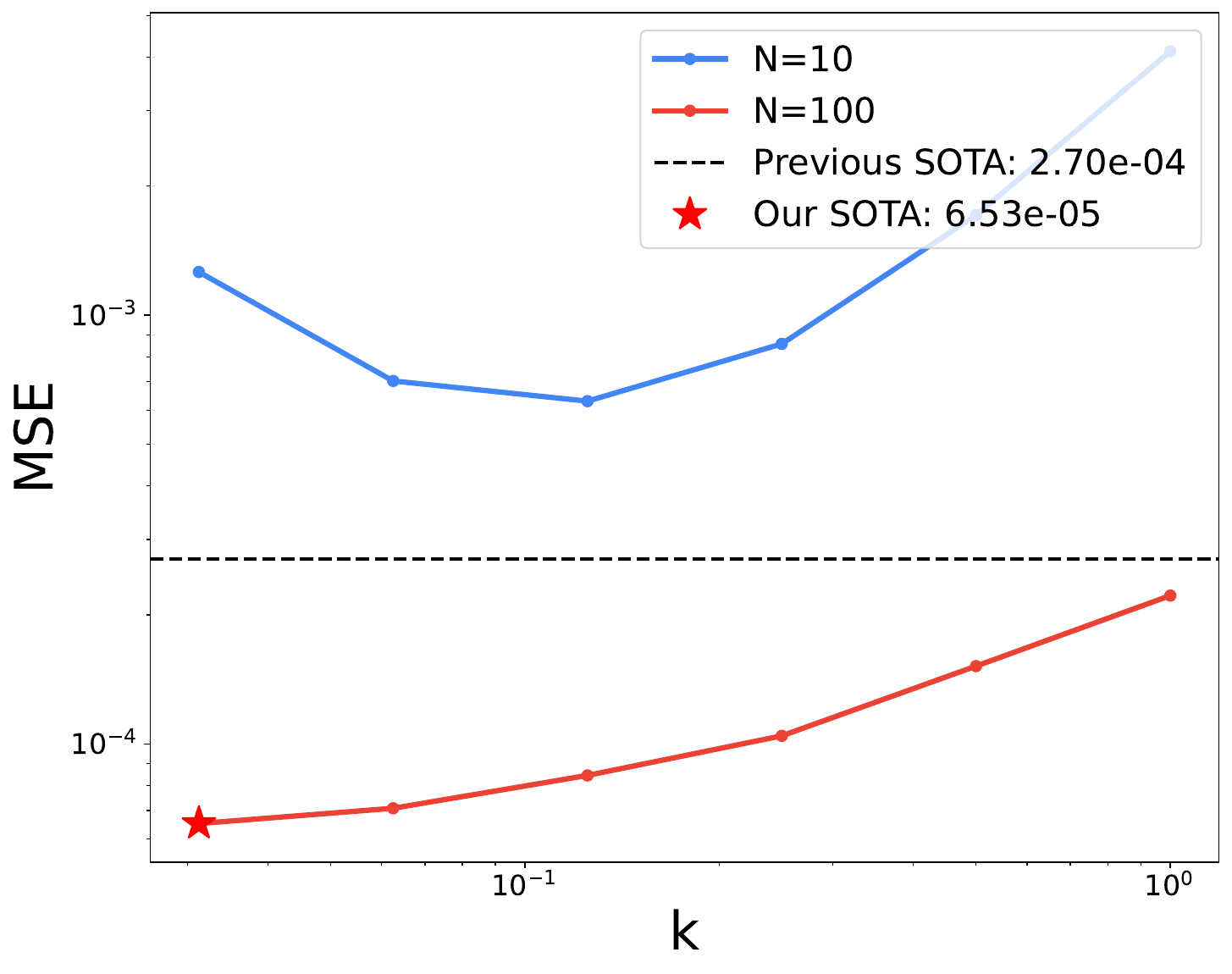}
    }
    \hfill
    \subfloat[MSE with changing $N$ and $k$ on Mujoco dataset imputation tasks, with missing ratio 0.8]{
        \includegraphics[width=0.3\textwidth]{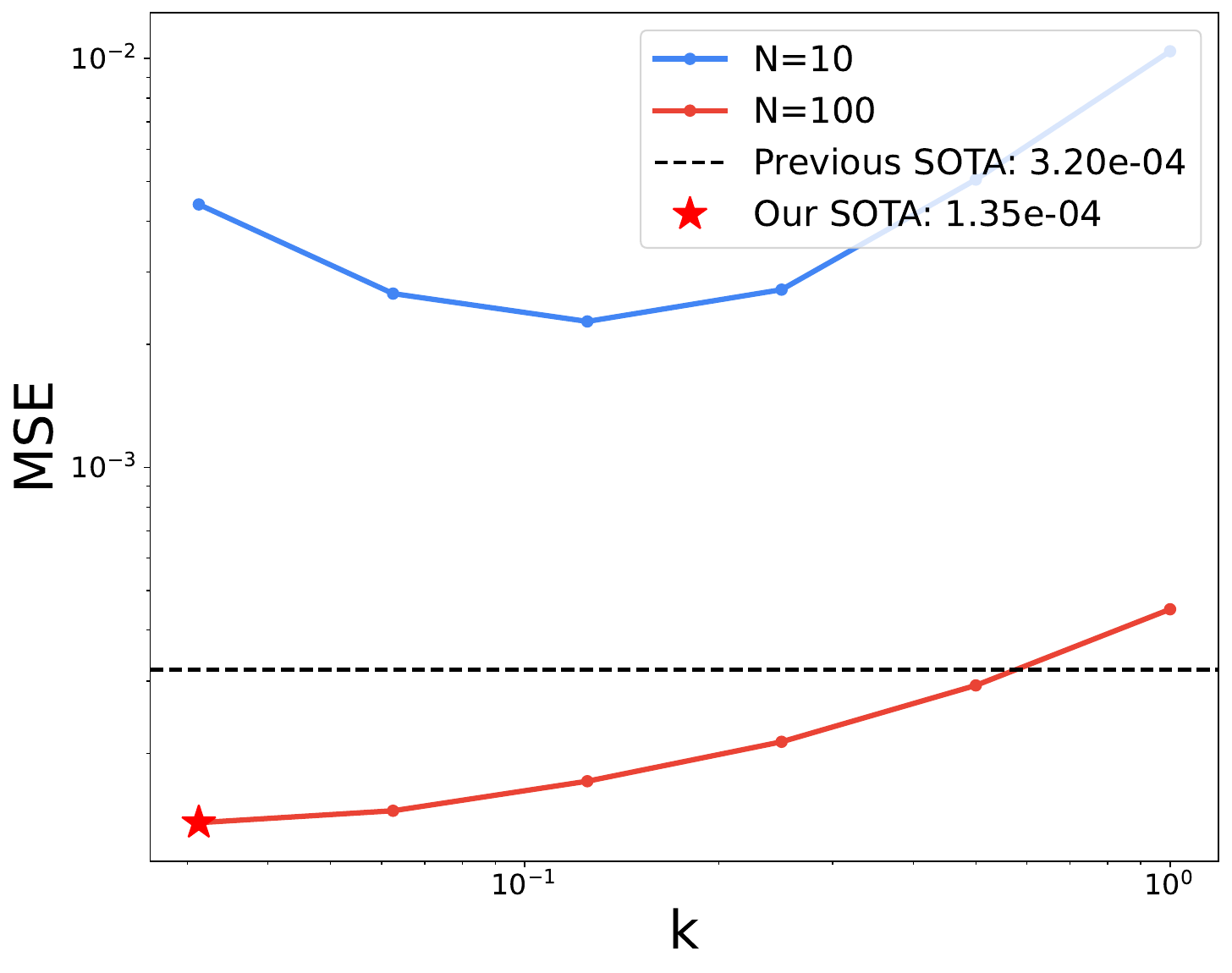}
    }
    \hfill
    \subfloat[MSE with changing $N$ and $k$ on solar dataset forecasting tasks]{
        \includegraphics[width=0.3\textwidth]{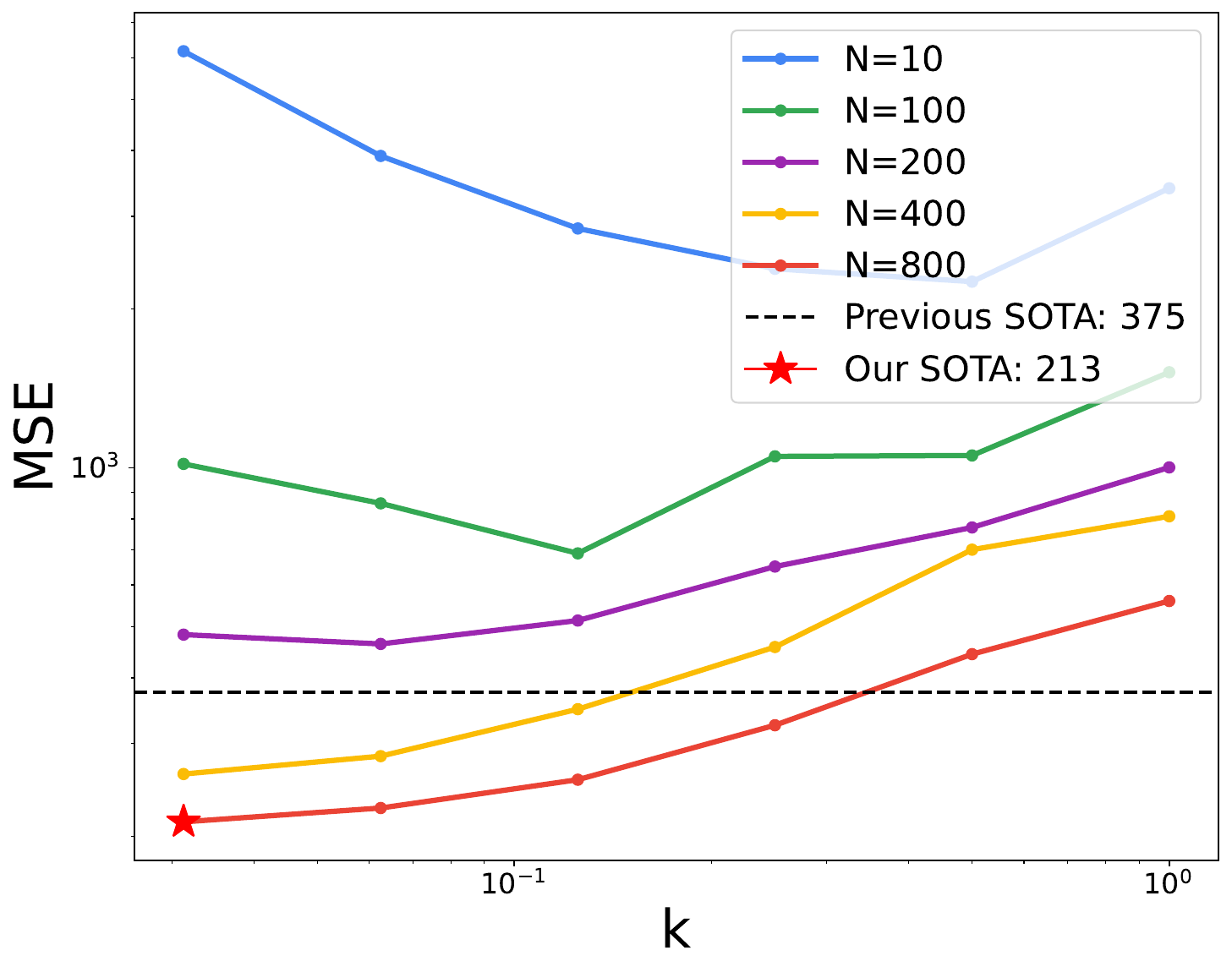}
    }
    \vspace{1em}
    \caption{Conditional generation with different $k$}
    \label{ksampling}
    \vspace{0mm}
\end{figure*}

\subsection{Further Experiments}
\label{sec:fur}
\textbf{Ablation Study}
Table~\ref{tab:ablation_study} shows that using both RoPE (frequency 50k) and attention registers (128 tokens) yields the lowest Context-FID score (0.1981 on ETTh, 0.1173 on Energy). Removing either component degrades performance: without both, FID scores rise sharply (0.9053 on ETTh, 0.3489 on Energy), while adding only RoPE offers partial improvement (0.2053 on ETTh, 0.3367 on Energy). This highlights the complementary benefits of positional encoding and global context tokens.

\begin{table}[htb]
    \centering
    \caption{Ablation study of FID (lower is better) on ETTh and Energy, showing that combining RoPE (frequency 50k) and attention registers (128 tokens) achieves the best scores. Base model represents the model without Rope and Register.}
    \label{tab:ablation_study}
    \renewcommand{\arraystretch}{1.1} 
    \resizebox{1\linewidth}{!}{ 
    \begin{tabular}{lcc}
        \toprule
        \textbf{Configuration} & \textbf{ETTh} & \textbf{Energy} \\
        \midrule
        Base model & 0.9053 $\pm$ 0.1611 & 0.3489 $\pm$ 0.0586 \\
        +Rope & 0.2053 $\pm$ 0.0191 & 0.3367 $\pm$ 0.0352 \\
        +Rope +Register & \textbf{0.1981 $\pm$ 0.0073} & \textbf{0.1173 $\pm$ 0.0134} \\
        \bottomrule
    \end{tabular}%
    }
\end{table}

\begin{figure}
    \centering
    \begin{minipage}[b]{0.5\textwidth}
        \centering
        \begin{subfigure}[b]{0.31\textwidth}
            \centering
            \includegraphics[width=\textwidth, height=0.75\textwidth, keepaspectratio]{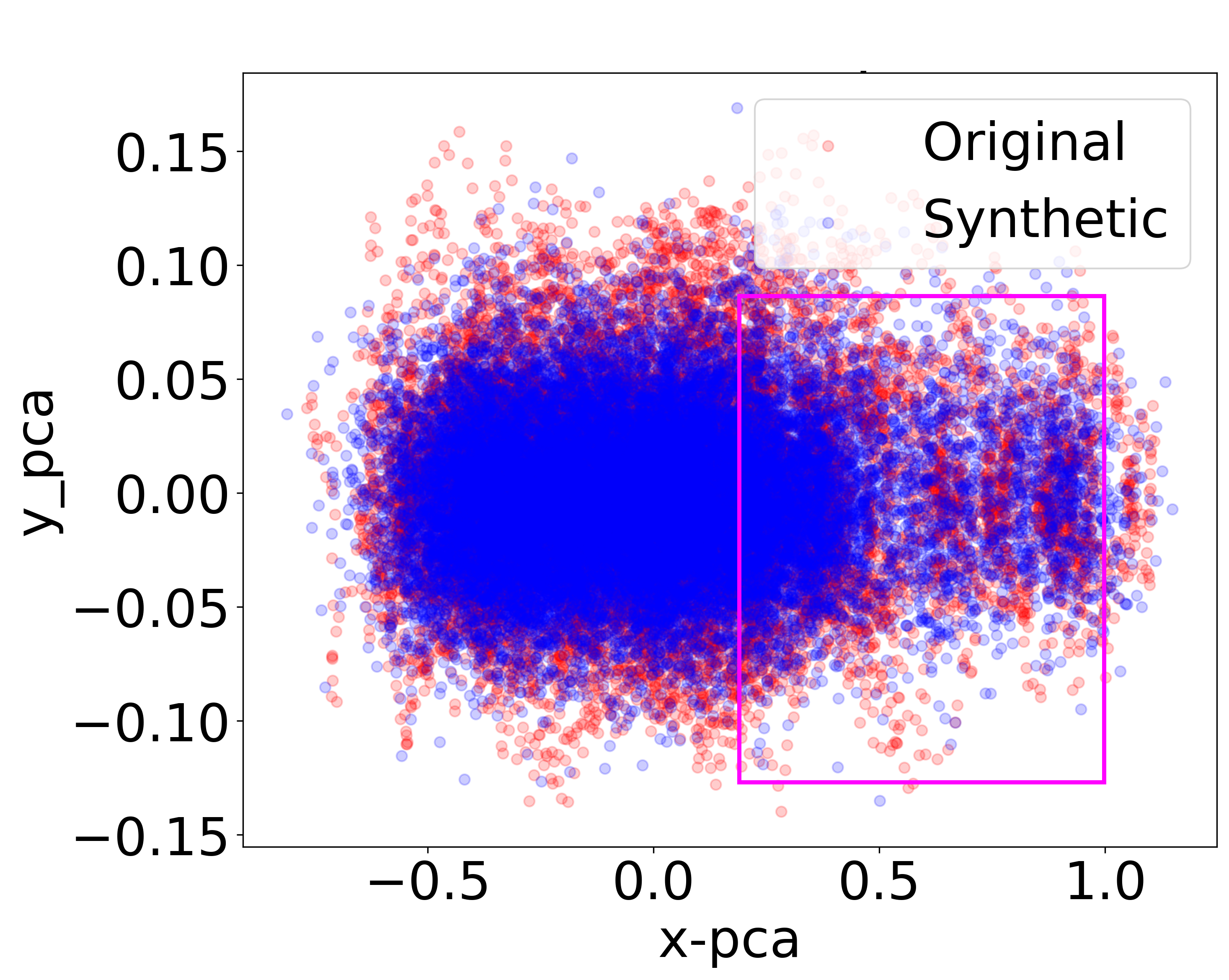}
            \caption{PCA (\textsc{FlowTS})}
            \label{fig:pca-fm-ts}
        \end{subfigure}
        \hfill
        \begin{subfigure}[b]{0.31\textwidth}
            \centering
            \includegraphics[width=\textwidth, height=0.75\textwidth, keepaspectratio]{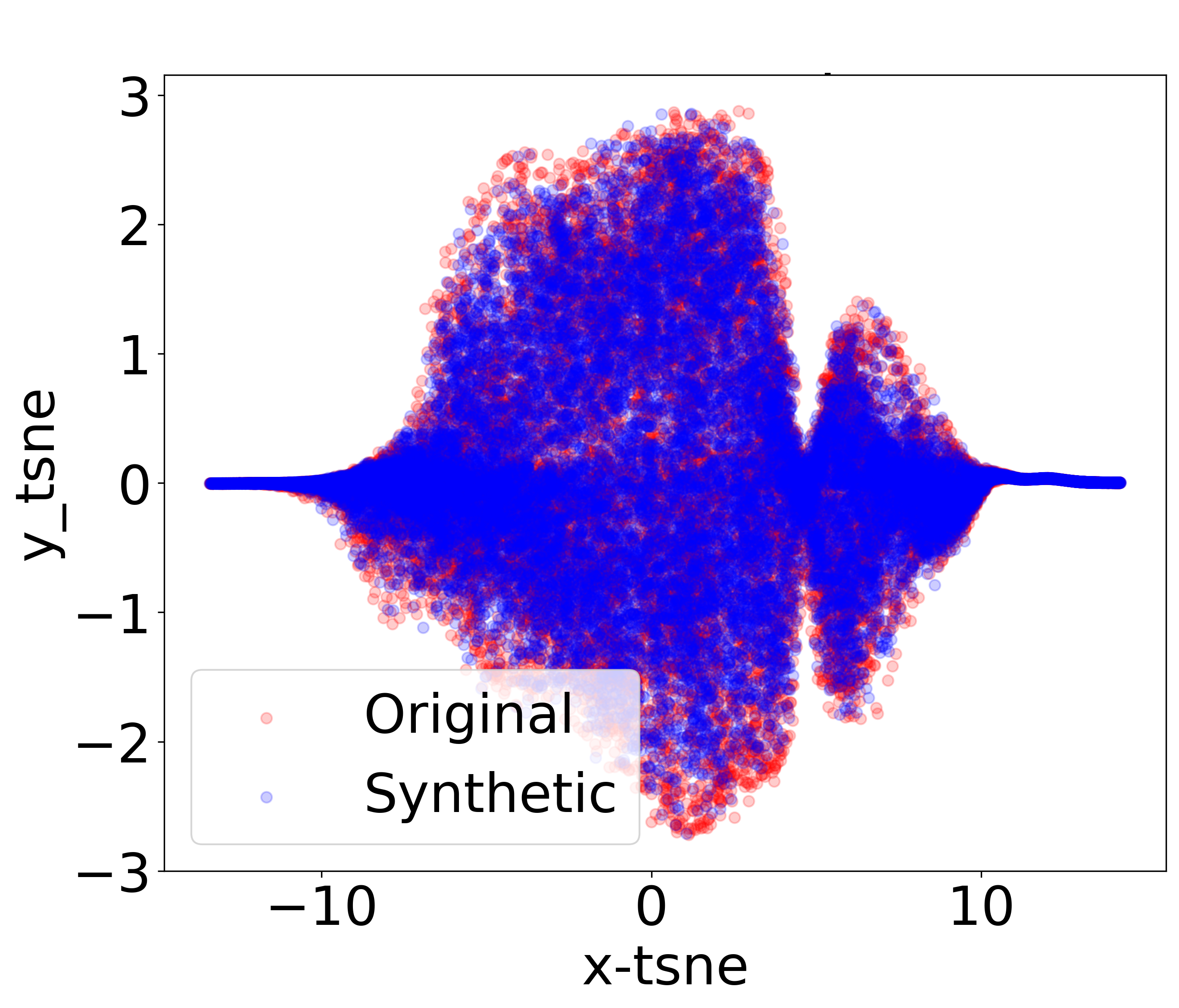}
            \caption{t-SNE (\textsc{FlowTS})}
            \label{fig:tsne-fm-ts}
        \end{subfigure}
        \hfill
        \begin{subfigure}[b]{0.31\textwidth}
            \centering
            \includegraphics[width=\textwidth, height=0.75\textwidth, keepaspectratio]{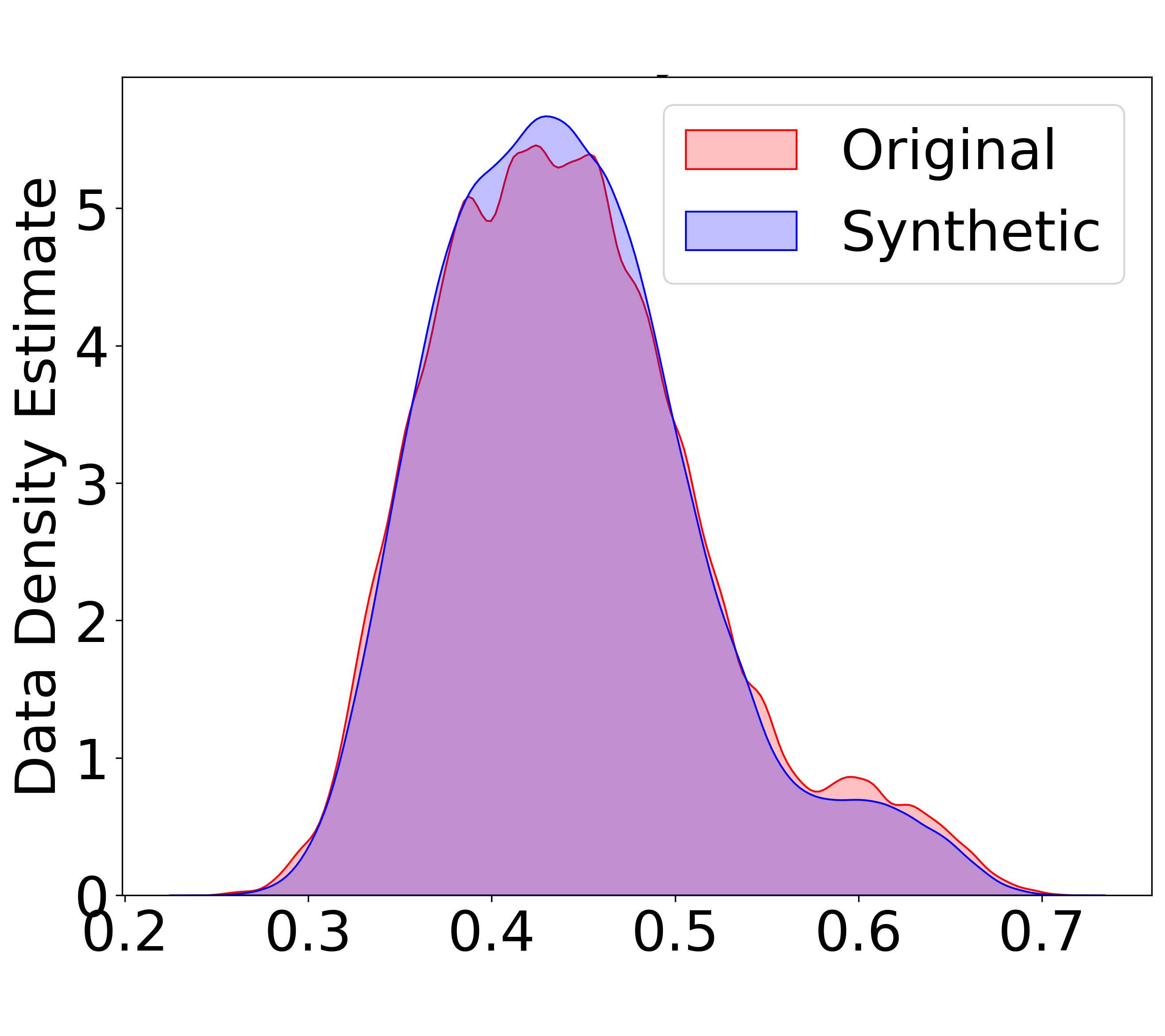}
            \caption{KDE (\textsc{FlowTS})}
            \label{fig:kde-fm-ts}
        \end{subfigure}
    \end{minipage}
    \begin{minipage}[b]{0.5\textwidth}
        \centering
        \begin{subfigure}[b]{0.31\textwidth}
            \centering
            \includegraphics[width=\textwidth, height=0.75\textwidth, keepaspectratio]{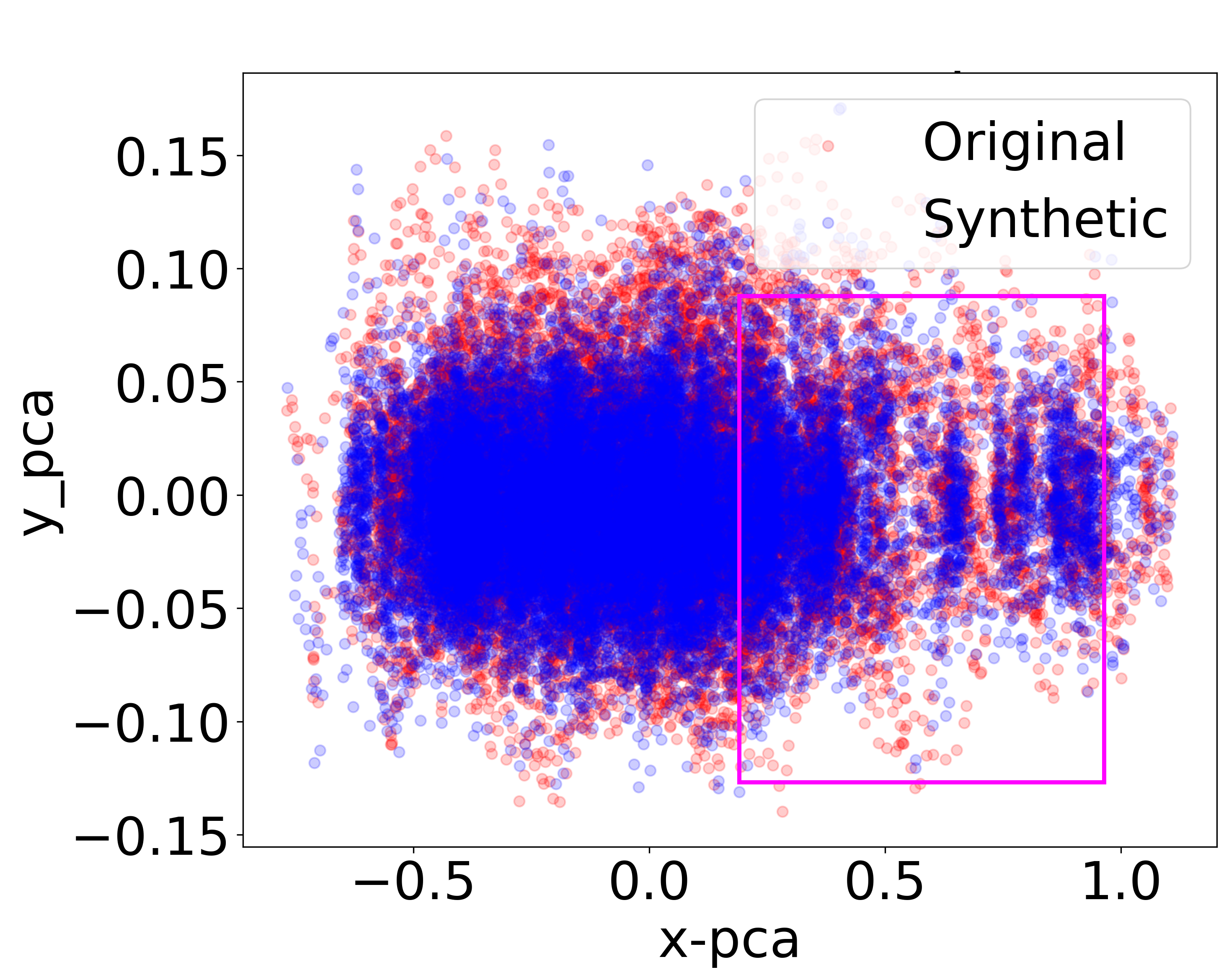}
            \caption{PCA (Diffusion-TS)}
            \label{fig:pca-diffusion-ts}
        \end{subfigure}
        \hfill
        \begin{subfigure}[b]{0.31\textwidth}
            \centering
            \includegraphics[width=\textwidth, height=0.75\textwidth, keepaspectratio]{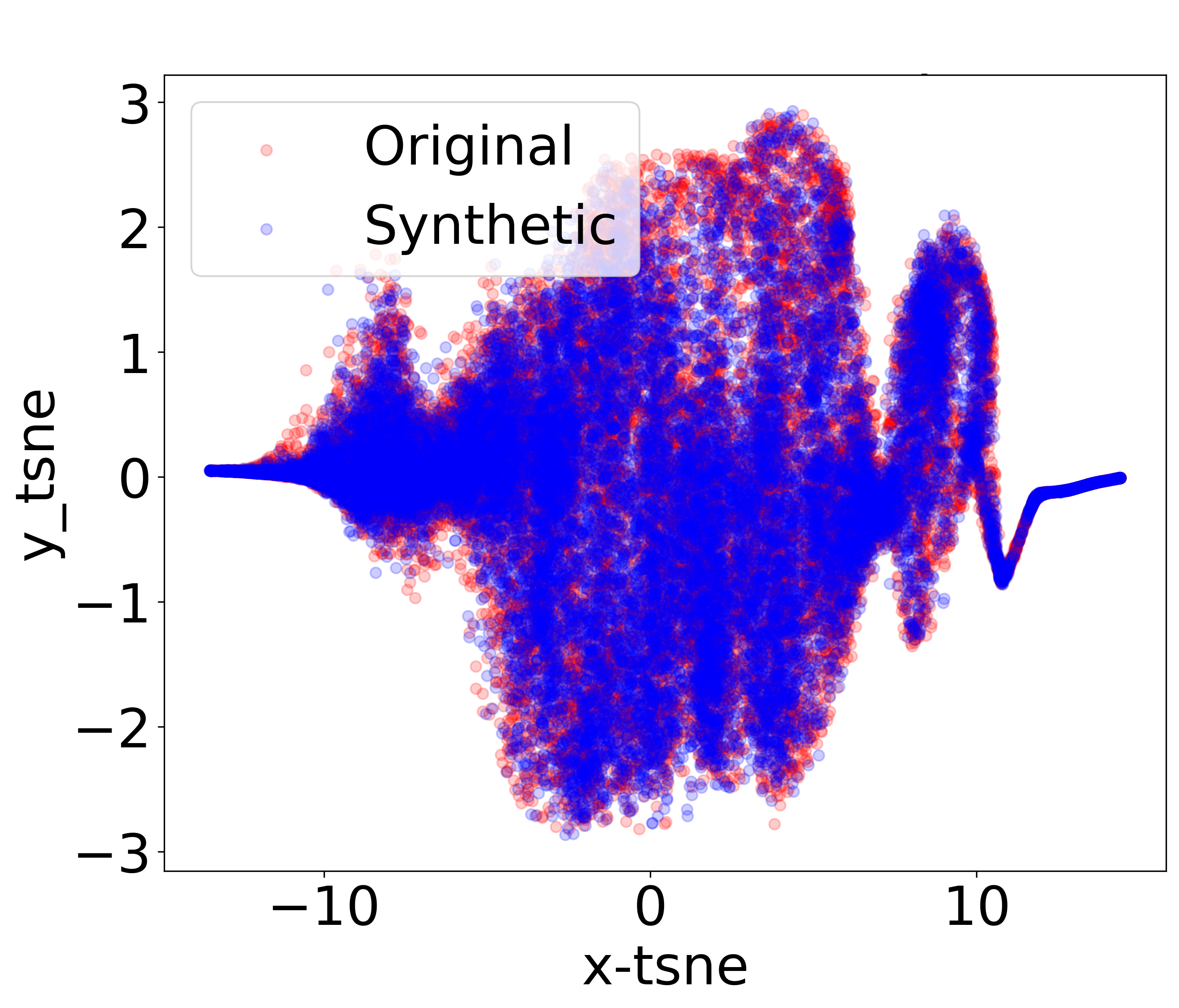}
            \caption{t-SNE (Diffusion-TS)}
            \label{fig:tsne-diffusion-ts}
        \end{subfigure}
        \hfill
        \begin{subfigure}[b]{0.31\textwidth}
            \centering
            \includegraphics[width=\textwidth, height=0.75\textwidth, keepaspectratio]{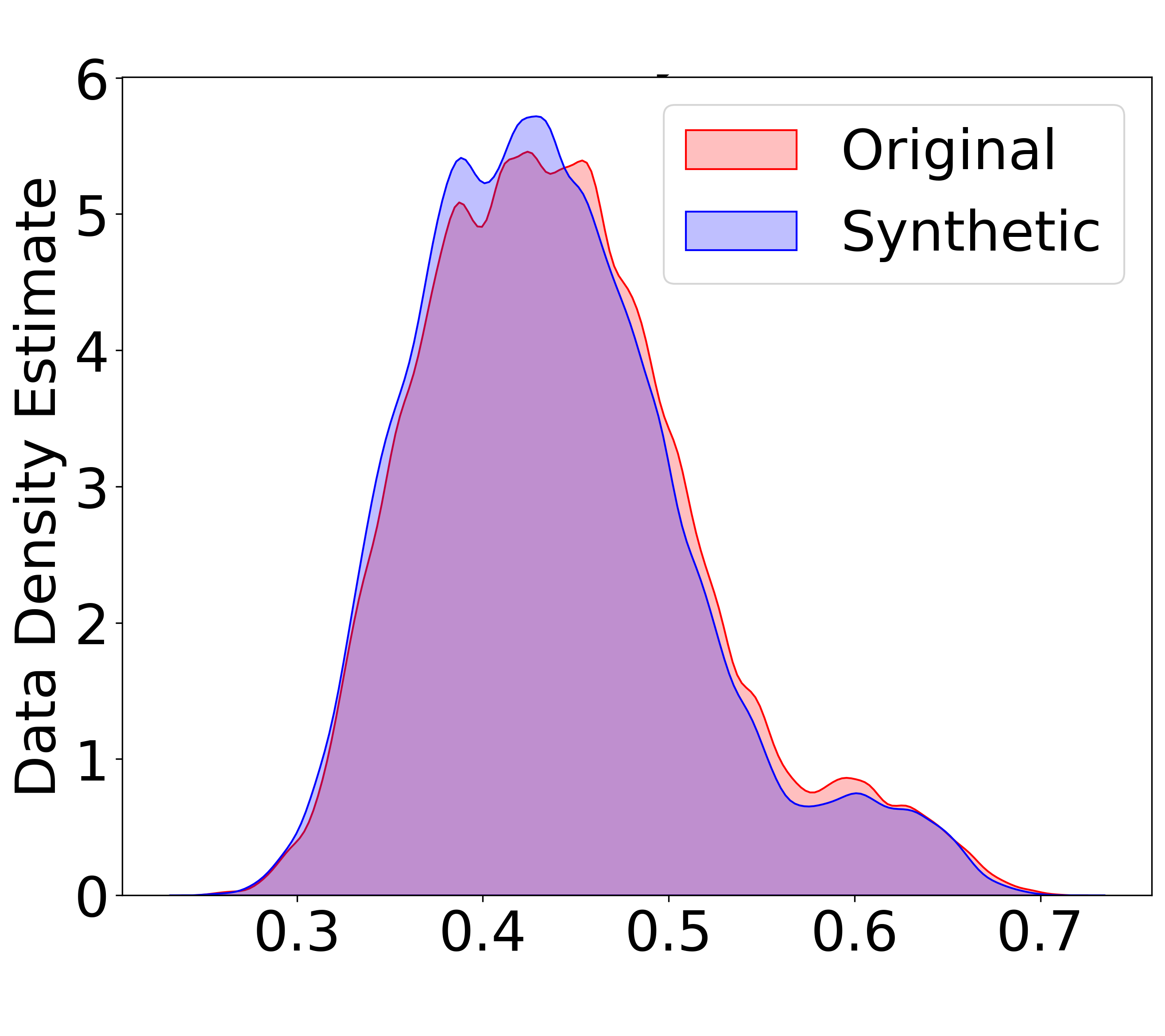}
            \caption{KDE (Diffusion-TS)}
            \label{fig:kde-diffusion-ts}
        \end{subfigure}
    \end{minipage}
    \caption{Embedding visualization comparison of generated sequences by \textsc{FlowTS} and Diffusion-TS methods relative to the target sequences using PCA, t-SNE, and Kernel Density Estimation. Here red indicates the target sequences, where blue indicates the generated sequences.}
    \label{tsne}
    \vspace{-1em}
\end{figure}

\textbf{\textbf{$k$}-selection}
In Fig.~\ref{ksampling}, we compare different values of $k \in (0,1]$ for generation under varying numbers of sampling iterations $N_s$. We evaluate the imputation task on Mujoco and the forecasting task on Solar with missing ratios of 0.7 and 0.8.
First, we demonstrate that adaptive sampling consistently achieves superior results across all settings. Second, we observe that as $k$ decreases, the MSE loss also decreases, indicating more stable inference. This suggests that a smaller $k$ can lead to better performance.



\textbf{Visualization}
To directly compare generated and target sequences, we follow the approach in~\citep{yuan2024diffusionts}, mapping both into an embedding space using PCA~\citep{shlens2014tutorialprincipalcomponentanalysis} and t-SNE~\citep{van2008visualizing}. 
As shown in Figure~\ref{tsne}, the KDE for \textsc{FlowTS} aligns more closely with the target sequences, especially on the right slope, where Diffusion-TS exhibits noticeable fluctuations, further validating \textsc{FlowTS}’s superior accuracy.


\section{{Conclusion}}
In this work, we address the critical computational inefficiency of diffusion-based time series generation by proposing \textsc{FlowTS}, an ODE-based model that replaces iterative ODE/SDE solvers with straight-line transport in probability space. 
By leveraging rectified flow and geodesic paths between two distributions, \textsc{FlowTS} achieves theoretical optimality while eliminating the need for costly iterative steps, significantly accelerating both training and inference. Our adaptive sampling strategy further enhances efficiency by dynamically balancing noise adaptation and precision.

To ensure high-quality generation, \textsc{FlowTS} integrates explicit trend and seasonality decomposition, attention registers for global context aggregation, and Rotary Position Embedding (RoPE) to encode positional dependencies. These components collectively address local and global temporal dynamics, enabling authentic and coherent time series synthesis. Extensive experiments validate \textsc{FlowTS}’s superiority, achieving state-of-the-art FID scores of 0.019 (Stock) and 0.011 (ETTh) and reducing solar forecasting MSE by 43.2\% compared to prior methods.

\newpage
\bibliography{refs}
\bibliographystyle{icml2025}

\newpage

\end{document}